\documentclass{article}

\usepackage[preprint]{neurips_2024}

\usepackage[utf8]{inputenc} %
\usepackage[T1]{fontenc}    %
\usepackage[frozencache,cachedir=.]{minted}
\usepackage{url}            %
\usepackage{booktabs}       %
\usepackage{amsfonts}       %
\usepackage{nicefrac}       %
\usepackage{microtype}      %
\usepackage{xcolor}         %
\usepackage{xspace}
\usepackage{listings}

\usepackage{wrapfig}
\usepackage{graphicx}
\usepackage{caption}
\usepackage{subcaption}
\usepackage{makecell}
\usepackage{soul}
\usepackage{tcolorbox}
\usepackage{hyperref}       %

\hypersetup{
    colorlinks=true,
    linkcolor=blue,
    filecolor=magenta,      
    urlcolor=magenta,
    pdftitle={Overleaf Example},
    pdfpagemode=FullScreen,
    }
    
\definecolor{codegreen}{rgb}{0,0.6,0}
\definecolor{codegray}{rgb}{0.5,0.5,0.5}
\definecolor{codepurple}{rgb}{0.58,0,0.82}
\definecolor{backcolour}{rgb}{0.98,0.98,0.97}

\definecolor{actionbg}{RGB}{101,165,66}
\definecolor{resbq}{RGB}{72,156,208}
\definecolor{querybg}{RGB}{218,120,66}

\DeclareRobustCommand{\hlquerybg}[1]{{\color{white}\sethlcolor{querybg}\hl{#1}}}

\DeclareRobustCommand{\hlresbg}[1]{{\color{white}\sethlcolor{resbq}\hl{#1}}}

\DeclareRobustCommand{\hlactionbg}[1]{{\color{white}\sethlcolor{actionbg}\hl{#1}}}

\lstdefinestyle{mystyle}{
    backgroundcolor=\color{backcolour},   
    commentstyle=\color{codegreen},
    keywordstyle=\color{magenta},
    numberstyle=\tiny\color{codegray},
    stringstyle=\color{codepurple},
    basicstyle=\ttfamily\footnotesize,
    breakatwhitespace=false,         
    breaklines=true,                 
    captionpos=b,                    
    keepspaces=true,                 
    numbersep=5pt,                  
    showspaces=false,                
    showstringspaces=false,
    showtabs=false,                  
    tabsize=2,
    frame=single,
}
\lstdefinelanguage{XML}
{
  morestring=[s]{>}{<},
  morecomment=[s]{<?}{?>},
  stringstyle=\color{black},
  identifierstyle=\color{black},
  keywordstyle=\color{magenta},
  morekeywords={thought,type,content}%
}

\newcommand{\model}{\mbox{\textsc{{CodeNav}}}\xspace}
\newcommand{\mnm}{\mbox{\textit{{m\&m's}}}\xspace}
\newcommand{\mthreeeval}{\mbox{\textsc{{M$^3$ToolEval}}}\xspace}
\newcommand{\apibank}{\mbox{\textsc{{API-Bank}}}\xspace}
\newcommand{\phidata}{\mbox{\textsc{{PhiData}}}\xspace}
\newcommand{\transformers}{\mbox{\mintinline{python}{transformers}}\xspace}
\newcommand\code[1]{\mbox{\mintinline{python}{#1}}}
\newcommand{\eg}{\emph{e.g.}\xspace}
\newcommand{\etc}{\emph{etc.}\xspace}
\newcommand{\ie}{\emph{i.e.}\xspace}

\title{\model: Beyond tool-use to using real-world codebases with LLM agents}

\author{%
  Tanmay Gupta* \quad Luca Weihs* \quad Aniruddha Kembhavi \\
  PRIOR @ Allen Institute for AI \\
  \url{https://codenav.allenai.org} \\
  \emph{* equal contribution}\\
}

\begin{document}

\maketitle

\begin{abstract}
  We present CodeNav, an LLM agent that navigates and leverages previously unseen code repositories to solve user queries. In contrast to tool-use LLM agents that require "registration" of all relevant tools via manual descriptions within the LLM context, CodeNav automatically indexes and searches over code blocks in the target codebase, finds relevant code snippets, imports them, and uses them to iteratively generate a solution with execution feedback. To highlight the core-capabilities of CodeNav, we first showcase three case studies where we use CodeNav for solving complex user queries using three diverse codebases. Next, on three benchmarks, we quantitatively compare the effectiveness of code-use (which only has access to the target codebase) to tool-use (which has privileged access to all tool names and descriptions). Finally, we study the effect of varying kinds of tool and library descriptions on code-use performance, as well as investigate the advantage of the agent seeing source code as opposed to natural descriptions of code. All code will be made open source under a permissive license.
  
\end{abstract}

\section{Introduction}\label{sec:intro}

Today, tool-use is the de facto approach for enabling LLMs to interact with external systems or programs to complete domain-specific tasks~\citep{Gupta2022Visprog,Suris2023ViperGPT,Wu2023AutoGen,Yang2023MMReact,Wang2023Voyager}.
In this tool-use paradigm, a list of functions, or more generally code snippets, are first ``registered'' with the LLM by adding their descriptions, potentially with examples of their usage, to the LLM's context. Then, given a user query, the LLM generates invocations of these tools, executed in an external environment, to solve the user's task.

While tool-use enables new capabilities, it also limits the expressiveness of LLMs by constraining them to invoke only a handful of meticulously described functions or API calls. With LLMs becoming increasingly capable at code understanding and generation~\citep{Wang2024Codeact,Chen2021Evaluating,GitHubCopilot2024,Jimenez2024SWEBench}, we argue that it is time to
move beyond \emph{tool-use} to \emph{code-use}, \ie move from meticulously designed and registered tools to a setting where the LLM agent directly reads, imports, and uses the source code of any given repository\footnote{We use the terms codebase, library, and repository interchangeably.}
to solve a user's query.

An effective code-use agent must be able to identify and use the right code snippets (functions, classes, constants, \etc) from the codebase to solve the given query.
This free-form search over code is made possible due to a simple observation: well-designed libraries are written by humans for humans.
These libraries often make crucial domain-specific assumptions, use meaningful abstractions and variable names, and organize code in files and directories so that it is easy to discover relevant functionality and
document
concisely. Instead of describing every single function or class in every file as done in tool-use, code-use leverages this structure %
to search for the required snippets.
Code-use may further benefit from a high-level library description that exposes the structure inherent in the codebase. This may be done in various ways: \eg, highlighting important directories, files, or entry points, describing the contents and purpose of complex files, or elucidating the library's abstractions and assumptions. This library description often already exists in the form of a README file, can be manually provided, or generated automatically via rule based parsing or an LLM.

\begin{figure}[t]
\centering
\includegraphics[width=0.8\textwidth]{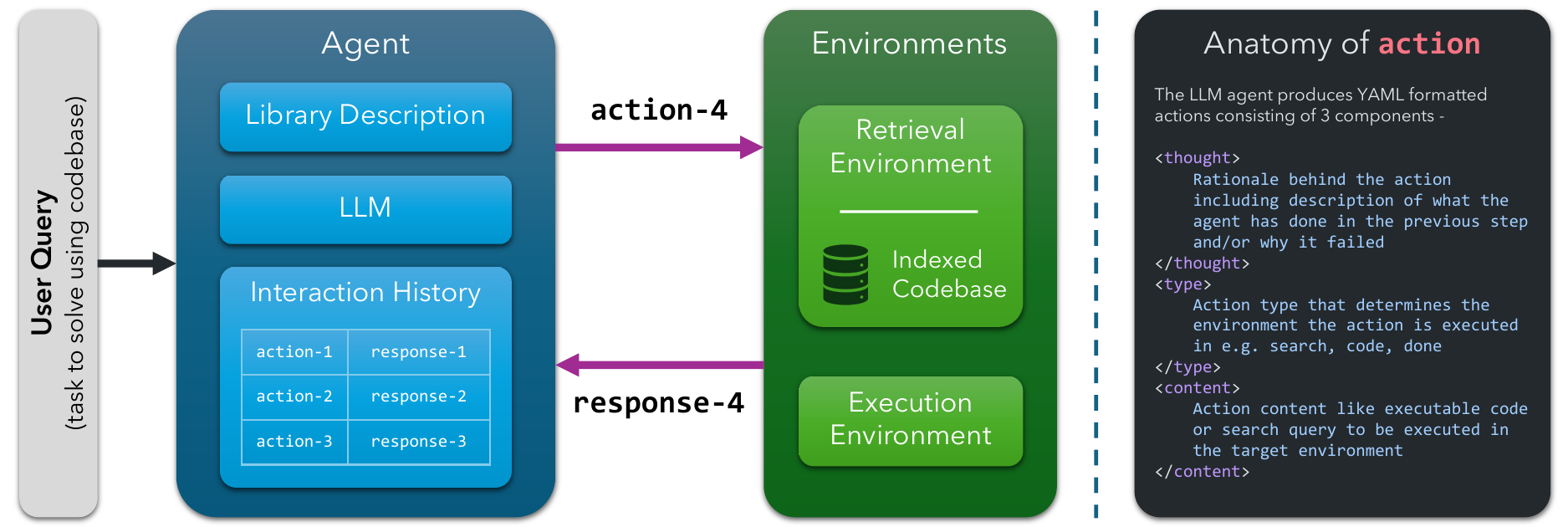}
    \caption{\textbf{\model{}'s single-agent, multi-environment interaction protocol.} Given a user query, a brief description of the codebase (\emph{library description}), and the interaction history, the LLM agent produces an \code{action} comprising of a \emph{thought}, action \emph{type}, and action \emph{content}. The action gets executed in the target environment (identified by action \emph{type}) to produce a \code{response}. The interaction at the current step consisting of the \code{action}-\code{response} pair is appended to the interaction history as context for the LLM to produce the next action.}
    \label{fig:teaser}

\end{figure}

We propose \model, a single-agent, multi-environment, interaction framework (see Fig.~\ref{fig:teaser}) where an agent \textsc{Nav}igates through a given \textsc{Code}base to find the code snippets it needs to solve a users' query. Given a user query and high-level library description, \model iterates between searching the codebase for useful code snippets and generating part of the solution code that imports, instantiates, or calls the relevant variables, classes, or functions defined in the retrieved results. In this iterative process, the agent inspects execution results, fixes errors (if any), searches to resolve ambiguities, and gradually builds a solution to the query.
The multi-environment setup makes it easy to extend the system's capability by adding a new environment (\eg, a \code{terminal}) along with its supported actions (\eg, \code{bash} commands) and its responses to those actions (\eg, \code{STDOUT}).

For evaluating \model, we begin by quantifying the gap between tool-use and code-use on three existing tool-use benchmarks: \mnm~\citep{Ma2024mnm}, \mthreeeval~\citep{Wang2024Codeact}, and \apibank~\citep{Li2023ApiBank}. To adapt these benchmarks for code-use, we  provide the files containing the tool implementations as the target codebase.
Tool-use forms an upper bound for code-use as the tool-use agent is privy to comprehensive, hand-crafted information that is available for these benchmarks.
Surprisingly, despite not having access to this privileged information, we find that \model is competitive with tool-use across these evaluations.
While these results are impressive, 
they fall short of demonstrating the full potential of code-use. To this end, we present three qualitative case studies using distinct codebases. In these case studies, we find that \model can follow complex multi-step queries, recover from execution errors, run iterative searches to better understand usage of code snippets, and format results as per user instructions to visualize computations (\eg, as a webpage).

We argue that code-use is more than just a niche application of generating code conditioned on another repository. We envision a future where code-use is how LLMs discover and use domain-specific tools without imposing any constraints on how the tools are developed or written (\eg, as a list of simplistic functions) and without any interventions (\eg, manual description and registration of tools) required for LLMs to use them. Our results are suggestive of a future where \textit{a codebase is all you need}!
We highlight six key contributions. (1) We introduce a novel code-use paradigm for LLM agents to move beyond tool-use to directly using real-world code bases to solve complex user queries. (2) We propose \model that formulates code-use as a multi-step interaction between a single LLM agent and stateful retrieval, and code execution environments. (3) On three tool-use benchmarks (\mnm, \mthreeeval, and \apibank), we find a minimal gap from code-use to the tool-use upper bound without requiring arduous
tool registration. (4) We study the effect of library or tool description richness on code-use performance. (5) We investigate the advantage of having access to the source code as part of retrieval result as opposed to just function signatures or docstrings. (6) We present three case studies to demonstrate the promise of code-use agents on solving complex queries using real-world codebases. 
Our code will be made open source under a permissive license.

\section{Related Work}

\textbf{Tool-use with LLMs.} Tool-use refers to an LLM identifying and invoking appropriate tools or functions to solve a task described by the user. Training-free methods for tool-use require first "registering" tools with the LLM. Tool registration techniques include \emph{descriptive} registration by describing the tools (\eg, signature and docstrings of the class or function that implements a tool) \citep{Hsieh2023ToolDoc,Suris2023ViperGPT}  or \emph{prescriptive} registration by providing in-context examples of similar task descriptions and corresponding tool invocations \citep{Gupta2022Visprog}. Training-based methods involve finetuning an LLM on instruction and tool invocation pairs \citep{Qin2023ToolLLM}. In this work, we focus on training-free methods for tool-use and code-use.

\textbf{Retrieval for tool-use.} Scaling training-free methods to a large number of tools is challenging due to limited (albeit steadily increasing) context windows of LLMs. An approach to circumvent this is retrieving only the relevant tool descriptions or usage examples from a library. For instance, \citet{Hsieh2023ToolDoc} employ TF-IDF search to retrieve relevant tool documentation. DocPrompting~\citep{zhou23docprompting} explores sparse and dense documentation retrieval for code generation. ART~\citep{Paranjape2023ART} use cross-task demonstration retrieval for multi-step reasoning and tool-use. EcoAssistant~\citep{Zhang2023EcoAssistant} saves past successful solutions to user queries in a database and for new queries retrieves solutions to similar queries in the database as demonstrations. Unlike previous works that retrieve documentation, our \model agent directly retrieves source code using boolean \texttt{Elasticsearch} queries (\eg \texttt{(type:} \texttt{CLASS)} \texttt{AND} \texttt{(text:} \texttt{ObjectDetection)}).

\textbf{Prompting strategies for tool-use.} For user queries that require multi-step tool-use solutions, the agent has to simultaneously exercise the ability to plan (which tools to use and when) as well as invoke the tools correctly. To improve planning, prompting strategies like Chain of Thought~\citep{Wei2022ChainOfThought} prompt the LLM to precede the actual solution with a \emph{thought} that describes the step-by-step plan.  ReAct~\citep{Yao2023ReAct} interleaves \emph{thought}, \emph{action} (tool invocation), and \emph{observation} (result of executing the action or feedback). CodeAct~\citep{Wang2024Codeact} generalizes ReAct by allowing \emph{actions} to be free-form code. In contrast to CodeAct, \model uses a single-agent, multi-environment framework in which the agent can both search for and execute code. Further, CodeAct operates in the tool-use regime where the exact tools that are needed for the query are registered ahead of time. %

\textbf{Code generation.} While not all tool-use methods require the LLM to generate executable code~\citep{Khot2022DecompPrompt} (instead generating function names and arguments and using a custom interpreter), tool-use may be implemented as free-form code generation \citep{Wang2024Codeact,Ma2024mnm}. Code generation with LLMs has been explored for a wide range of applications including code-completion in code editors~\citep{GitHubCopilot2024}, generating functions from docstrings~\citep{Chen2021HumanEval}, editing files in a repository to fix Github issues~\citep{Jimenez2024SWEBench}, and even generating an entire repository consisting of multiple files from a single natural language instruction~\citep{gptengineer}.

\textbf{Feedback and correction.} As tasks get more complex, a single LLM call may only produce a partial or partially correct solution. For such tasks, \emph{agentic workflows} enable a closed-loop system~\citep{Wu2023AutoGen,Wang2023Voyager} where the LLM agent iteratively produces an intermediate solution, receives feedback on the solution, and proceeds to either fix any errors or generate the next step. In context of code generation and tool-use, the feedback may consist of execution output such as variable values and exceptions raised, output of test cases~\citep{Huang2023AgentCoder}, output of static analysis tools like linting and type-checking, and human or LLM feedback~\citep{Madaan2023SelfRefine}. Recent works have also explored emulating execution feedback using an LLM~\citep{Li2023ChainOfCode,Ni2024NExT}.

\section{\model}\label{sec:model}

\subsection{Overview}

We formulate code-use with LLMs in a single-agent, multi-environment interaction framework consisting of stateful code retrieval and code execution environments (see Fig.~\ref{fig:teaser}). The agent is given a brief high-level library description (\eg important directory or file paths, description of content and purpose of complex files, crucial abstractions and assumptions made by the library, \etc) and the user query to be solved. The agent then proceeds to interact with these environments over multiple rounds. Each interaction consists of an \code{action} from the agent and a \code{response} from an environment. %
Each \code{action} consists of a (i) \emph{thought} used for chain-of-thought reasoning~\citep{Wei2022ChainOfThought,Yao2023ReAct}, (ii) an \emph{action type} specifying the environment the agent wishes to act upon (\eg code, search, \etc), and (iii) the \emph{action content} which is executed in the selected environment. The action content gets routed to the appropriate environment based on the action type and the environment executes the content updating its state and producing a \code{response}. The history of past interactions is provided to the agent as context to generate the next action. The interactions continue up to a maximum number of interactions specified by the user or until the agent takes the \code{done} action. We now describe details of environments, actions, and responses.

\subsection{Environments}

For code-use, the agent must be able to perform two essential functions: (i) search for or discover relevant code snippets in the target codebase; (ii) generate the next portion of the code solution that imports, instantiates, or calls the needed classes or functions. In \model, the agent performs these functions by taking actions in one of the following environments. %

\textbf{Retrieval Environment}. This environment serves as an interface to a search index for fetching code snippets from the target codebase with rule-based re-ranking and a persistent memory to avoid resurfacing past retrievals. We parse the entire codebase and index all functions, classes, import statements, assignments, {\etc} as individual documents. We implement the index using \code{Elasticsearch} (\code{ES}) with fields for code string, code type (function, class, assignment, import etc.), file path, and line numbers~\citep{elasticsearch2024}. The agent can use a \code{search} action type to issue search queries into this index. Given the action, the environment first uses the action content as the search query to the \code{ES} index, discards any matches that have already been retrieved by past searches during the episode, and then re-ranks the results based on heuristic rules (\eg, to prioritize functions and classes and de-prioritize assignments and import statements). Finally, the top-$k$ results are added to the environments persistent memory and returned as the environment reponse. Issuing the same action again surfaces the $k$ next-best matches.  

\textbf{Execution Environment}. Actions of type \code{code} get routed to a \code{Python} execution environment.\footnote{In this work we consider only Python code generation for simplicity but here is no limitation on \model preventing extension to other languages.} At the beginning of the episode, the environment is initialized with an empty \code{global_variables} dictionary. Each code block is executed in scope of these global variables (using \code{exec(code_str, global_vars)}) and any changes to the global namespace (\ie, modification or deletion of existing variables, or creation of new variables) are reflected in this dictionary. Prior to execution, the environment optionally performs linting (using \code{flake8}), type-checking (using \code{mypy}), and formatting (using \code{black}) of the code block. Standard output (\code{stdout}), updated variables in \code{global_vars} (new variables or variables whose string representations have changed), and errors if any (execution, linting, or type-checking) are returned as part of the response.   

\textbf{Done and Code Summary Environments}. Additionally, we create a few helper environments. The \emph{Done Environment} returns a \code{null} response to the \code{done} action that marks the end of the episode. The \emph{Code Summary Environment} treats the content of the \code{code_summary} action as a cleaned up summary of the code solution produced by the agent during the episode and saves it for easy access.

\subsection{Agent Actions}

The \model\ agent interacts with the environments using an \code{action} that consists of 3 components: \emph{thought}, \emph{type}, and \emph{content}, see  Fig.~\ref{fig:teaser} (\emph{right}). The LLM underlying the agent is prompted to only produce outputs in a structured XML format and in compliance with a set of rules (\eg, \emph{thought} and \emph{type} must always be provided; \emph{type} must be one of the available action types; \etc). Before routing the \code{action} to the target environment, we check its validity. If the \code{action} is found to be invalid, an \code{InvalidAction} response is returned to the agent containing a description of the violated rule. The agent may use this violation description to fix the \code{action} in the next step.

\begin{figure}[hbtp]
\centering
\includegraphics[width=1\textwidth]{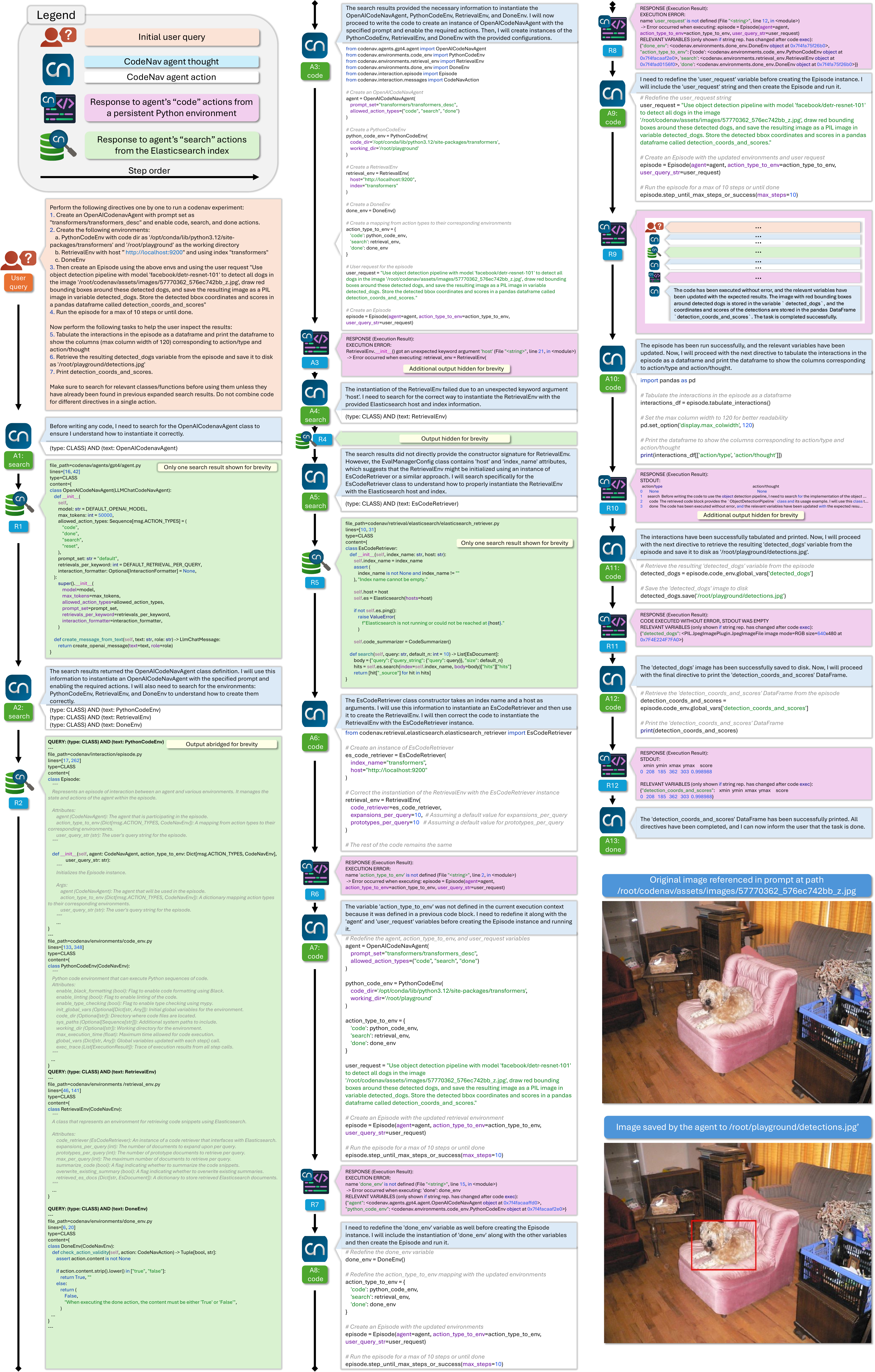}
    \caption{\textbf{Running the \model{} agent with the \model{} codebase.}}
    \label{fig:codenav-on-codenav}
\end{figure}

\subsection{Environment Responses}
Each action elicits a response from the target environment or an \code{InvalidAction} response if the action is invalid for execution in the target environment. Each response in \model\ is implemented as a data class with a \code{format()} method that specifies how the response data should be serialized to a text string to be included in the agent's context when predicting the next action.

\textbf{Retrieval Response}. Given a search query (\ie, the \emph{content} of a \code{search} action), the retrieval environment returns a list of documents (containing code blocks from the target codebase along with metadata) from the search index that match the search queries. Implementation of the \code{format()} method of the retrieval response answers the question: what should be shown to the agent from these retrieved code blocks? On one hand, the agent may gain a better understanding of how to use a class or a function by reading its source code (containing function signatures, argument types, outputs, and implementation details) as opposed to reading an imprecise, incomplete, outdated, or entirely absent human written description or \code{docstring} of the class or function. On the other hand, showing all implementation details for every retrieved code block results in an explosion in the number of context tokens to be processed by the LLM. To strike a balance, we first retrieve up to $M$ ($=100$) matched documents. From these, we show the top-$K$ ($=3$) matches with source code and metadata. For large codebases, this is usually insufficient to surface target code blocks unless the exact function or class name is given in the search query. Therefore, we additionally show \emph{prototypes} (signatures and filenames) for up to $P$ classes or functions in the remaining retrieved results. Further, for the top-K matches, we use GPT-4 to generate \code{docstrings} for the top-3 retrievals and show whichever is shorter between the source code and the function signature with the generated docstring. %

\textbf{Execution Response}. Given a \code{code} action, the Python execution environment executes the \emph{content} producing a response. The \code{format()} method of a response serializes standard output (\code{stdout}), variables changed during execution shown as variable names along with string representation of their values, execution errors if any, and (optionally) linting, type-checking, and formatting errors. The error messages contain reference to the line in the code string that produced the error to help localize the error. The \code{stdout} and changed variables allow easy inspection of function calls but can get quite long (\eg, when printing a large array) and are therefore truncated to a maximum number of characters. We show the start and end of \code{stdout} and the beginning of variable values.

\section{Case Studies}\label{sec:case-studies}

While we quantitatively evaluate \model on tool-use benchmarks in Sec.~\ref{sec:exp}, these benchmarks are not sufficiently complex to highlight the advantages of code-use over tool-use. Therefore, we showcase \model's impressive capabilities in three case studies using diverse codebases to solve complex queries. %
For the first case study (Sec.~\ref{sec:codenav_case_study}), Fig.~\ref{fig:codenav-on-codenav} depicts the entire episode. For the other two case studies, we show the inputs and outputs in Fig.~\ref{fig:case_studies} and provide the full episodes as part of the supplementary material. For all case studies we provide library descriptions in the appendix (App.~\ref{app:prompts}). Please also see App{.} Table~\ref{tab:codebase-stats} for information about the size and complexity for the codebases used in these case studies and our quantitative experiments (\eg, searching the \texttt{transformers} library requires searching over 50{,}508 snippets in 3{,}475 files).

\subsection{\model{} on \model{}}\label{sec:codenav_case_study}

We imagine a researcher who, possibly after reading this paper, wishes to use \model to answer a query using the \texttt{transformers} library~\citep{wolf2020transformers}. In place of a researcher however, we use a \model{} agent; \ie, a \model agent uses the \model repository to instantiate another \model agent to answer a given query with \texttt{transformers}.
This example serves two goals: (1) it provides a pedagogical example of using our codebase, and (2) 
it shows \model's zero-shot abilities as we can guarantee that the underlying LLM (GPT-4) was not trained on our codebase.

For this case-study, our user query consists of 7 steps divided into 2 distinct parts (see \hlquerybg{User~query} in Fig.~\ref{fig:codenav-on-codenav}). Steps 1-4 specify instructions for creating and running the episode while Steps 5-7 contain instructions to visualize the results of the interaction. 
In \ul{Steps 1-4}, the user asks \model\ to first create an agent using \code{OpenAICodenavAgent} and to instantiate various environments using \code{PythonCodeEnv}, \code{RetrievalEnv}, and \code{DoneEnv} with the specified parameters like the \texttt{Elasticsearch} host and index name to use for retrieval. Then the query asks the agent to create an episode for solving another query (\emph{within the original query!}) using \code{transformers}. This ``subquery'' requires the agent to detect dogs in an image (specified by a file path) using the \code{facebook/detr-resnet-101} model in the object detection pipeline, add red detection boxes on the image, and store the image in variable \code{detected_dogs}. The subquery also asks agent to store the detection coordinates and scores as a pandas dataframe
in the variable \code{detection_coords_and_scores}.
The first part of the full query ends with asking the agent to run the episode for a maximum of 10 steps
\ul{Steps 5-7} specify how to visualize the interaction. Specifically, we ask the agent to: (i) tabulate the interaction as a dataframe with columns for action type and thought; (ii) save the \code{detected_dogs} image as a PNG at a specified file path; and (iii) print \code{detection_coords_and_scores}.

The \model episode for the above can be found in Fig.~\ref{fig:codenav-on-codenav}. The agent initially searches for information about the \code{OpenAICodenavAgent} class and the environments (\hlactionbg{A1}, \hlresbg{R1}, \hlactionbg{A2}, \hlresbg{R2}), and then attempts to instantiate them with code (\hlactionbg{A3}). This code results in an error due to a misuse of the \code{RetrievalEnv} initializer (\hlresbg{R3}). The agent then searches for additional information to resolve this error (\hlactionbg{A4}, \hlresbg{R4}) and eventually succeeds in running a new \model agent using \texttt{transformers} (\hlactionbg{A9}, \hlresbg{R9}). Finally it prints and saves the requested outputs.

\begin{figure}[t!]
\centering
\includegraphics[width=0.9\textwidth]{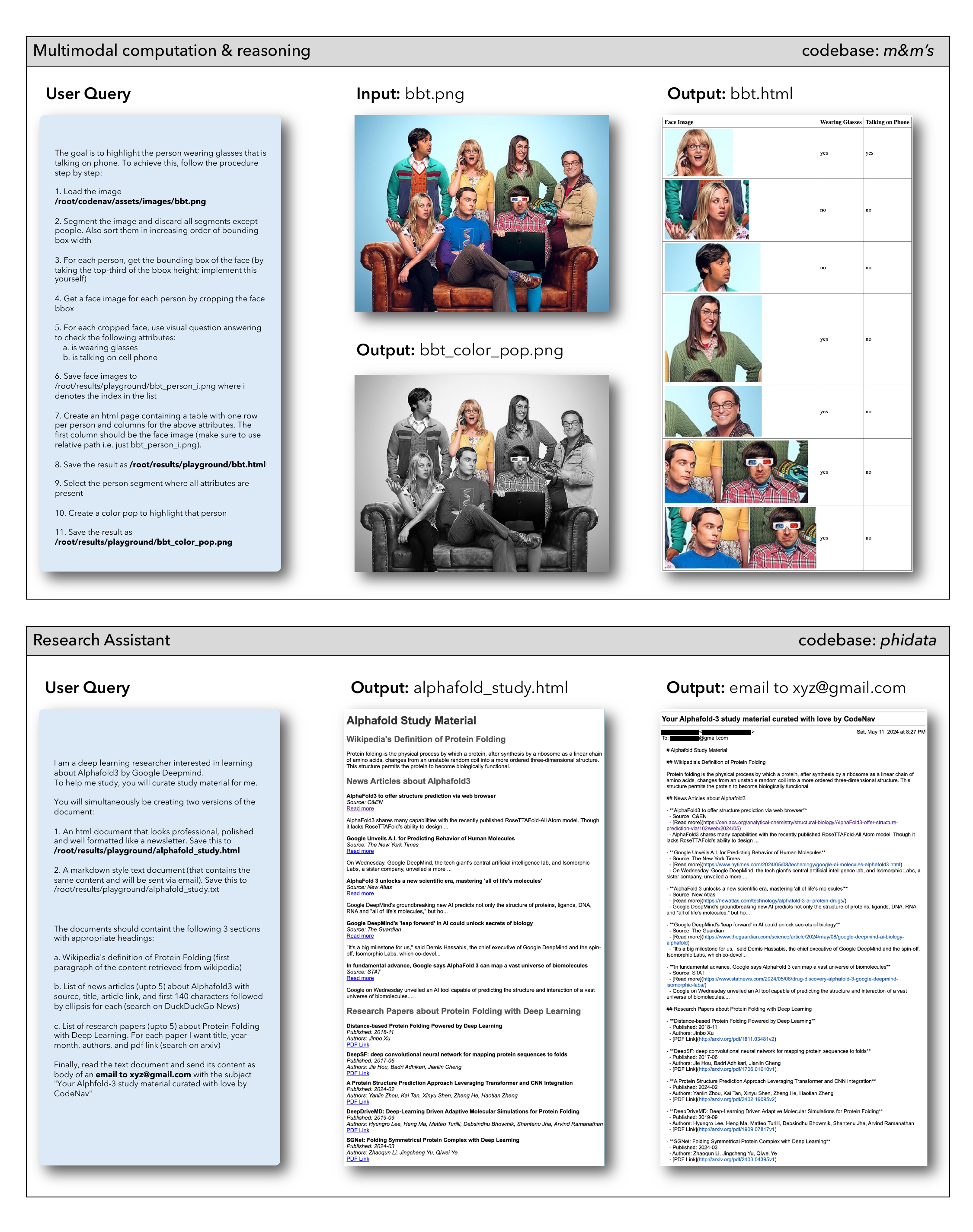}

\caption{\textbf{\model\ unifies "agentic" applications via code-use.} Two case studies: (\emph{top}) a visual reasoning and image editing agent; and (\emph{bottom}) an information gathering agent. These applications are enabled simply by changing the target codebase search index and the high-level library description.}
\label{fig:case_studies}

\end{figure}

\subsection{Multimodal Processing and Reasoning}\label{sec:mnm_case_study}

In our proposed code-use formulation, multimodal tasks are identical to text-only tasks so long as multimodal processing functionality is available in the target codebase. We demonstrate this with an image editing task requiring visual reasoning to localize the region to edit. We use the \mnm codebase as it contains computer vision tools for detection, segmentation, QA, \etc
In this case-study, see Fig.~\ref{fig:case_studies} (top), the agent is required to find and highlight \textit{the person wearing glasses who is talking on the phone}.
Our query first specifies steps to localize the region to edit. In particular, the agent is instructed to first segment the image and select the \textit{person} segments. Then for each person, the agent should zoom in on the face by taking a crop of the top-third of the person bounding box. For each face, the agent must use visual question answering to verify whether the person is wearing glasses \emph{and} is talking on cell phone. To visualize these predictions, the agent is instructed to save these face crops along with the predictions as an HTML table. Finally, the agent is instructed to highlight the person for which both attributes are true via a color-pop effect.
We observe that \model not only uses multimodal models proficiently but also uses the outputs to perform visual reasoning. Further, this case study highlights \model's ability to produce human-interpretable intermediate outputs.

\subsection{Research assistant}\label{sec:phidata_case_study}

Imagine an agent that curates reading material on a given topic including news articles, blog posts, and research papers and then emails it to you. For \model, this is simply a matter of using a codebase that provides the necessary functionality for querying knowledge sources on the web and sending emails. One such codebase is \phidata~\citep{phidata}. 
In this case-study (see Fig.~\ref{fig:case_studies}, bottom) we query the agent to curate reading material on "Alphafold-3" consisting of definition from Wikipedia, a list of news articles, and a list of papers on "Protein folding with Deep Learning". Further, our query specifies various presentation requirements: \eg, for each news article, we require the agent to show the article source, title, link, and first 140 characters.
Finally, we ask the agent to write this information to an HTML and markdown-formatted documents. The text document's content is then sent to the user's email address with the specified subject.
This case study demonstrates the versatility of \model to create specialized agents simply by providing an appropriate codebase. Here \model serves as a research assistant by using functionality built in \phidata to query Wikipedia, DuckDuckGo News, and arXiv.

\section{Experiments}\label{sec:exp}

We now present our quantitative results on 3 tool-use benchmarks: (1) \mnm, which requires multi-step planning with 33 multi-modal, \eg vision/language, tools~\citep{Ma2024mnm}; (2) \apibank that involves managing user state in sandboxed environment via calls to any of 73 APIs~\citep{Li2023ApiBank}; and (3) \mthreeeval that contains ``82 human-curated tasks'' requiring multiple tools, calls, and interactions~\citep{Wang2024Codeact}. In all experiments we report the mean performance across 3 independent evaluations; the reported $\pm$ values correspond to $2{\times}$ the standard dev{.} across evaluations. As these benchmarks were not necessarily designed for evaluation with code-use agent in mind (\eg, \apibank uses JSON API calls), this has necessitated some changes in how we evaluate \model on these benchmarks. See App.~\ref{app:exp-details} for these details as well as descriptions of our metrics.

\subsection{How does code-use compare to tool-use on tool-use benchmarks?}

Tool-use benchmarks are designed to test LLMs' ability to invoke a small set of pre-registered tools. Since tools in these benchmarks are relatively simple function calls with human written descriptions, code-use is upper bounded by tool-use on these benchmarks. We wish to quantify the gap between code-use 
(where source code search is necessary) 
and tool-use
(where no search is needed as tool names/descriptions are provided).
Tab.~\ref{tab:tool_vs_code} shows that on \mnm\ and \apibank, code-use achieves similar or slightly lower tool-f1. On both benchmarks, code-use takes a minor hit on tool-recall. This is intuitive as tool-use is provided tool names
while code-use has the harder task of searching to discover available tools.
On \mthreeeval, which evaluates final answer correctness, code-use is within two points of tool-use. In all datasets on average, code-use takes ${\sim}$2 more interaction steps compared to tool-use to search for required tools. Finally, code-use results in only a minor increase in performance variance despite the added uncertainly due to lack of knowledge of available tools.

\begin{table}[]
\caption{\textbf{Code-use is competitive with tool-use even without tool prompts.} 
}
\resizebox{\textwidth}{!}{%
\begin{tabular}{lcccccccccccc}
\toprule
 & \multicolumn{4}{c}{\mnm}                                                                                            & & \multicolumn{2}{c}{\mthreeeval}             &        & \multicolumn{4}{c}{\apibank}                                                                                             \\
\cmidrule(r){2-5}  \cmidrule(r){7-8} \cmidrule(r){10-13} 
\textbf{method}    & \textbf{precision}                     & \textbf{recall}                        & \textbf{f1}                            & \textbf{steps}       &          & \textbf{accuracy}                      & \textbf{steps}         &        & \textbf{precision} & \textbf{recall}    & \textbf{f1}        & \textbf{steps} \\ \midrule
tool-use  & 82.9 \textpm\ 4.5 & 81.7 \textpm\ 0.4 & 79.6 \textpm\ 2.3 & 4.9 \textpm\ 0.1 & & 83.7 \textpm\ 2.8 & 6.6 \textpm\ 0.5 & & 86.6 \textpm\ 0.8 & 93.6 \textpm\ 1.1 & 88.5 \textpm\ 0.7 & 3.4 \textpm\ 0.1      \\
code-use  & 88.0 \textpm\ 6.1 & 78.2 \textpm\ 4.5 & 80.6 \textpm\ 5.1 & 7.2 \textpm\ 0.2 & & 81.7 \textpm\ 4.9 & 7.8 \textpm\ 0.4 & & 84.0 \textpm\ 0.7 & 89.3 \textpm\ 0.6 & 85.3 \textpm\ 0.3 & 5.3 \textpm\ 0.0   \\
\bottomrule
\end{tabular}%
}
\label{tab:tool_vs_code}
\vspace{-0.5em}
\end{table}

\subsection{Is a library description sufficient for tool-use?}

\begin{wraptable}{r}{0.4\textwidth}
\vspace{-3em}
\caption{\textbf{Desc{.} ablations on
\mnm.}}
\vspace{-0.5em}
\centering
\label{tab:tool_desc}
\resizebox{0.4\textwidth}{!}{%
\begin{tabular}{lrcc}
\toprule
\textbf{tool description}   & \textbf{length}       & \textbf{f1}           & \textbf{steps}       \\ \midrule
w/o desc & 0 & 74.1 \textpm\ 1.9 & 7.0 \textpm\ 0.1 \\
tool names     & 694            & 78.1 \textpm\ 3.5 & 6.7 \textpm\ 0.2 \\
\quad + desc     & 3680         &  80.8 \textpm\ 0.4 & 6.9 \textpm\ 0.1 \\
\qquad + prototypes & 4627 &  80.7 \textpm\ 5.0 & 6.1 \textpm\ 0.1 \\ \midrule
\makecell[l]{library desc \\(\model)}  & 2061      & 80.6 \textpm\ 5.1 & 7.2 \textpm\ 0.2 \\
\bottomrule
\end{tabular}%
}
\end{wraptable}

Instead of meticulously listing each tool or function name along with detailed descriptions of its purpose and input and output arguments, \model\ allows a user to only provide a high-level description of the codebase or library that implements these tools. In Tab.~\ref{tab:tool_desc}, we compare library description (Fig.~\ref{fig:lib_desc_mnm} in appendix) with \model\ without any description as well as tool descriptions with three "levels of detail"; the lowest level contains only the tool names while the richest contains tool names, descriptions, and function signatures (Fig.~\ref{fig:tool_desc_mnm} in appendix). As providing tool details in the prompt helps the agent identify the tools needed to solve a query and use them correctly, we find a consistent increase in tool-f1 with increasing tool detail. Spectacularly, library description achieves similar performance as the richest tool description with less than half the description length. The convenience of not requiring description of each tool comes at the cost of minor increase in number of steps since now the agent needs to search and discover the tool as opposed to recalling from context.

\subsection{Does seeing the source code help code-use?}

\begin{wraptable}{r}{0.35\textwidth}
\vspace{-3em}
\caption{\textbf{Search response formatting ablation on \mnm.}
} %
\vspace{-0.5em}
\centering
\label{tab:retr_response}
\resizebox{0.35\textwidth}{!}{%
\begin{tabular}{lcc}
        \toprule
        \textbf{retrieval response}                 & \textbf{f1}           & \textbf{steps}       \\ \midrule
        prototypes                & 80.0 \textpm\ 5.2 & 7.8 \textpm\ 0.1 \\
        code      &  81.2 \textpm\ 2.3 & 7.3 \textpm\ 0.2 \\ \midrule
        \makecell[l]{code or docstring \\ (\model)}  & 80.6 \textpm\ 5.1 & 7.2 \textpm\ 0.2 \\
        \bottomrule
        \end{tabular}%
}
\vspace{-0.5em}
\end{wraptable}

Well written code is its own documentation. Therefore, for an agent 
that is proficient in code understanding, seeing the actual implementation details in the source code (which might also include docstrings) alleviates the need to manually register tools and provides strictly more information than just the function signatures or prototypes.  %
We compare various retrieval response formats in Tab.~\ref{tab:retr_response} using the \mnm\ benchmark. We observe that returning code for the top-3 matches results in higher tool-f1 than showing prototypes only in the retrieval response. We also see a reduction in number of interaction steps needed by the agent to solve the query which has been a consistent indicator of lower uncertainty in deciding which tools to use and how. Finally, real-world code blocks can span 100s of lines which quickly increases the context length to be processed by the agent. To remedy this we generate docstrings for the top-3 retrievals using GPT-4 and between the docstring and the raw code we show whichever is shorter. As expected, this default configuration in \model\ achieves tool-f1 higher than prototypes only but lower than code.

\subsection{What makes a good library description?}

\begin{wraptable}{r}{0.4\textwidth}
\vspace{-3em} %
\caption{\textbf{Comparing two library descriptions on \mthreeeval.} To provide a reference for length, desc{.} length for tool-use for \mthreeeval\ is 5K.}

\centering
\label{tab:lib_desc}
\resizebox{0.4\textwidth}{!}{%
\begin{tabular}{lccc}
\toprule
\textbf{library description} & \textbf{length} & \textbf{accuracy} & \textbf{steps} \\ \midrule
file path               & 253     & 76.8 \textpm\ 6.5      & 8.3 \textpm\ 0.3    \\
file path + file desc   & 641     & 81.7 \textpm\ 4.9      & 7.8 \textpm\ 0.4 \\ \bottomrule
\end{tabular}%
}

\end{wraptable}

We demonstrate the impact of library description in Tab.~\ref{tab:lib_desc}. \mthreeeval is implemented as a codebase consisting of 5 files, each containing tools dedicated to a problem domain; web browsing, travel planning, dna sequencing, message encryption, and financial calculations. The first library description simply provides relative file paths to these files (\eg, \emph{m3eval/travel\_planner.py}), while the second description also includes a one line summary of what the file contains (\eg, \emph{``functions for planning travel including finding flights, making hotel reservation, and budget calculations''}). Intuitively, this enables \model\ to come up with keywords for search and results in superior performance with fewer interactions needed to reach a solution. We provide the following three recommendations to write good library descriptions: (i) provide context for the target domain to enable the LLM to then use its knowledge of the domain to generate useful keywords for search; (ii) describe library structure (\eg, directory structure or key assumptions the library makes); (iii) provide a brief (not necessarily exhaustive) natural language description of the available functionality.
\emph{}

\subsection{Impact of LLM choice on performance}

\begin{wraptable}{r}{0.5\textwidth}
    \vspace{-3em} %
    \centering
    \caption{\textbf{LLM choice ablation on \mnm.}}\label{tab:llm-ablation}
    
    \resizebox{0.5\textwidth}{!}{%
    \begin{tabular}{lccccc}
    \toprule
    \textbf{LLM}  & \textbf{precision}    & \textbf{recall}       & \textbf{f1}           & \textbf{steps}       \\ \midrule
    \texttt{gpt-4-1106-preview}  
    & 88.0 \textpm\ 6.1 
    & 78.2 \textpm\ 4.5 
    & 80.6 \textpm\ 5.1 
    & 7.2 \textpm\ 0.2 \\
    \texttt{gpt-3.5-turbo-0125}  
    & 54.36 \textpm\  2.3
    & 15.77 \textpm\  1.4
    & 22.96 \textpm\ 0.8
    & 9.08 \textpm\ 1.07 \\
    \texttt{Mixtral-8x22B-Instruct-v0.1}
    & 82.50 \textpm\  2.1
    & 62.31 \textpm\  1.9
    & 67.91 \textpm\ 1.3
    & 9.06 \textpm\ 0.3  \\
    \texttt{Qwen1.5-110B-Chat}  
    & 78.15 \textpm\  3.1
    & 38.49 \textpm\  5.5
    & 48.84 \textpm\ 5.1
    & 10.00 \textpm\ 0.2  \\
    \bottomrule
    \end{tabular}
    }%
    
\end{wraptable}

We have used GPT-4~\citep{OpenAI2023GPT4}, in particular \texttt{gpt-4-1106-preview},\footnote{\url{https://platform.openai.com/docs/models/gpt-4-turbo-and-gpt-4}} as the LLM underlying \model{}. As GPT-4 is one of the most performant publicly available LLMs, it is natural to ask how \model{} performance degrades when using smaller, possibly open source, LLMs.
We run \mnm evaluations using \model{}
when replacing GPT-4 with GPT-3.5, the Mistral 8$\times$22B mixture-of-experts model~\citep{Jiang2023Mistral,Jiang2024Mixtral}, and the Qwen1.5 110B model~\citep{Bai2023Qwen}.\footnote{Model completions obtained via the \texttt{together.ai} API, see \url{https://docs.together.ai}.} We chose to use the Mistral and Qwen models as they represent some of the largest, open-source, LLMs with long context windows (empirically, LLMs with context sizes below 16k tokens regularly fail by exceeding their context limit). Our results are displayed in Table~\ref{tab:llm-ablation}.
While the GPT-4 powered \model outperforms, the open source models do perform quite well falling behind primarily in their recall (suggesting search failures). Surprisingly the GPT-3.5 variant performs poorly; when inspecting the failed trajectories this appears to often be caused by the agent failing to appropriately summarize its actions at the end of the episode which may ameliorated by additional prompt tuning.

\section{Discussion}

We have argued that it is time to move from tool-use to code-use; from feeding LLM agents manually curated and meticuluously descibed tool sets, to instead presenting them with existing codebases written by humans for humans. As we have shown in our case-studies and quantitative results, simply remarkable behavior can be obtained by code-use agents when using modern LLMs so long as considerable care is taken to engineer code search and execution environments that provide the agent with significant flexibility and feedback.

{\small
\bibliographystyle{abbrvnat}
\bibliography{neurips_2024}
}

\newpage
\appendix
\section*{Appendices}

This appendices contain the following:
\begin{itemize}
    \item Experiment details for \mnm, \mthreeeval, and \apibank (App.~\ref{app:exp-details})
    \item Compute requirements (App.~\ref{app:compute})
    \item Library and tool descriptions (App.~\ref{app:prompts})
    \item A full retrieval response that shows code, automatically generated summary docstrings, and prototypes (App.~\ref{app:retr_response_example})
    \item Limitations of \model (App.~\ref{sec:limitations})
    \item Societal impact (App.~\ref{app:societal-impact})
\end{itemize}

Please see our project website, \url{https://codenav.allenai.org/}, for: 
\begin{itemize}
\item  Full episode trajectories for the 3 case studies.
\item An example run of \model on \mnm. This contains an HTML file that can be opened in a browser to view the programs generated by \model along with ground truth programs and links to trajectories.
\item Similarly, an example runs of \model on \mthreeeval with HTML visualizations.
\item A side-by-side comparison of library and tool descriptions for the \mnm and \mthreeeval codebases. 
\end{itemize}

\begin{wraptable}{r}{0.5\textwidth}
\vspace{-4em}
\centering
\caption{\textbf{Codebase statistics.}}
\vspace{-0.75em}
\label{tab:codebase-stats}
\resizebox{0.5\textwidth}{!}{%
\begin{tabular}{lrrrrrr}
\toprule
\textbf{Codebase}            & \textbf{files} & \textbf{snippets}  & \textbf{lines}   & \textbf{characters}    & \textbf{functions} & \textbf{classes} \\
\midrule
\mnm          & 2     & 58    & 971     & 34277    & 39        & 0       \\
\mthreeeval       & 8     & 39    & 485     & 14009    & 21        & 2       \\
\apibank     & 54    & 163   & 6144    & 207656   & 1         & 53      \\
\model      & 36    & 369   & 4055    & 141636   & 58        & 46      \\
\phidata      & 426   & 2549  & 49731   & 2169989  & 133       & 398     \\
\transformers & 3475  & 50508 & 1362242 & 63105978 & 4354      & 11424  \\
\bottomrule
\end{tabular}%
}
\vspace{-0.5em}
\end{wraptable}

\section{Experiment details} \label{app:exp-details}
For each of the 3 tool-use benchmarks used in \model evaluation, we now provide details on how the benchmark was adapted for code-use as well as metrics used.

\subsection{\mnm} %

\subsubsection{Adaptation for code-use}
All tools in \mnm are implemented in two files; a single python file with tool functions and a config file. Since functions in \mnm neither contain type hints nor are annotated with inline comments or docstrings, to provide minimal context, we add the tool description provided in the original \mnm benchmark as a single-line docstring to each function. For instance, for the function \code{color_pop}, we add \emph{It takes an image and one or multiple objects, and returns an image where only the object is colored and the rest is black and white}. Note that we provide no additional information about input arguments or outputs. Further, \mnm evaluation expects a sequence of function calls as outputs, we enable a `code\_summary` action in the agent to get the code solution in the desired format. To comply with the evaluation we provide the following guidelines:
\begin{verbatim}
When solving tasks YOU MUST RESPECT the following guidelines:
1. Do not implement any new functions. Just use the available functions.
2. Generally, try searching for function names. Only if needed, include 
function argument names. Do not include argument values.
3. When you have a solution, use the code_summary action to summarize 
the solution.
4. When asked to generate text, don't generate text yourself but rather 
see if there is a function to do it.
5. Tasks typically require 1 to 3 function calls.
\end{verbatim}

\subsubsection{Metrics}
We use the macro-averaged tool-f1 metric from the original \mnm paper~\citep{Ma2024mnm} which is the harmonic mean of the precision and recall of the function names in reference to the ground truth program. When multiple correct ground truths are available we used the best match. Since many queries in \mnm do not have a single correct answer, we do not use answer correctness as a metric. Similarly, we find that there are many ways of specifying the arguments while generating free-form code for \mnm queries and hence find argument name and value based metrics unreliable for evaluating free-form code generation agents like \model.

\subsubsection{Data split}
\mnm consists of approximately 800 samples. Since evaluating \model on the full set using GPT-4 as the LLM could cost between \$150 to \$200 (and we run each experiment thrice to compute error bars), we randomly sample a smaller set of 200 samples for our evaluation.

\subsection{\mthreeeval}
\subsubsection{Adaptation for code-use}
To prepare \mthreeeval for code-use, we begin by creating a codebase consisting of just the tool implementations and associated data (web page data used by their web browsing tools as well as flight, hotel, and location information used by the travel planning tools). Particularly for web browsing, while \mthreeeval registers methods of the \texttt{WebBrowser} class as individual tools, for code-use we let the agent use the class directly. Further, since the web browsing task uses unrealistic pages with strong assumptions about how the pages are formatted, we provide some context to the agent for the web browsing task as guidelines. Finally, tools in \mthreeeval are grouped by task (e.g. all tools for DNA sequencing are in the same file), and tasks only use tools from a single file. Therefore, to let the agent make use of this assumption, we provided file paths in library/tool description and we ask the agent to identify and specify the relevant file name in its search queries to zone in on required tools. Here are the guidelines we use:

\begin{verbatim}
When solving tasks YOU MUST RESPECT the following guidelines:
1. For browsing tasks use the WebBrowser object and navigate the web pages 
using available methods to find what you are looking for. Sometimes the 
relevant information may not be visible on the page but if you see 
[Viewing page m of n] (where m < n) then you may use the scroll functions 
to see more page content. If you see the information you need displayed on 
the web page, feel free to use it directly without worrying about parsing 
it using code. Do not write complex code. Rather, try to interact with the 
browser one action at a time like clicking or scrolling. 
2. You may need to identify the relevant python file and specify this target
file in your search queries to get the relevant search results
\end{verbatim}

\subsubsection{Metrics}
We use the final answer accuracy as used in the original \mthreeeval work~\citep{Wang2024Codeact}.

\subsection{\apibank}

\subsubsection{Adaptation for code-use}

The \apibank benchmark with a human-AI ``chat'' context in mind where an agent and user send messages back and forth with the user asking the agent to, possibly, perform many tasks one after another. The AI agent is then evaluated via a next-step prediction approach where the agent is fed the entire chat context up to time $t$ and required to predict some ground-truth chat message or JSON API call at time $t+1$. As \model was not designed to be used for back-and-forth chat with a user (neither is it meant to be evaluated on producing natural language responses) we filter all ground-truth interactions in the \apibank \texttt{level-1-given-desc} set to include only those chats for which the last user message is followed by at least one message from the agent where the agent invokes an API call. After filtering, we are left with 186 (of originally 214) samples. During evaluation, we then give our \model the full chat context up to and including the last user message (and also modify the sandbox to be in the state up to this point) and then evaluate \model's ability to produce all remaining API calls.

In order for \model to make API calls, we require that it directly instantiate the appropriate \apibank class and then invoke the \texttt{call} method on that class (\eg, the model might instantiate the \texttt{AddAgenda} class as \texttt{aa = AddAgenda()} and then run \texttt{aa.call(token, content, time, location)} with \texttt{token}, \texttt{content}, \texttt{time}, and \texttt{location} variables it has previously defined. In order to encourage this behavior, we include instructions of the form:
\begin{verbatim}
When solving tasks YOU MUST RESPECT the following guidelines:
1. When calling APIs you should instantiate the relevant class and use the 
`call` method defined in the class. DO NOT USE INVOKE OTHER METHODS ON 
THE CLASS, YOU MUST ONLY CALL THE `call` METHOD.
2. Everything can be solved by calling APIs, do not define new APIs or 
modify the existing ones.
\end{verbatim}
in the library description given to the \model agent.

Note that the above differs substantially from how the agents are traditionally evaluated with \apibank where they, generally, produce JSON formatted API calls which are routed by \apibank to the correct class and \texttt{call} method.

\subsubsection{Metrics}

As noted above, we evaluate \model's ability to produce the correct remaining API calls given some chat context. As for the \mnm benchmark evaluation, we only evaluate \model's ability to call the correct APIs and ignore, for ease of evaluation, whether these APIs were called with the correct arguments or produced the correct results. Supposing that \model called a sequence of APIs $A=\{a_1, ..., a_n\}$ and that the ground-truth set of APIs' called was $G=\{g_1,..., g_m\}$, we count the number of matches between $A$ and $G$ (counting multiplicities) and compute recall as $R=(\# matches) / |G|$ and precision as $P=(\# matches) / |A|$ (precision is taken to be 0 if $|A|=0$). Given this precision and recall, we compute the F1 score as usual as $F1 = 2\cdot P \cdot R / (R + P)$ with $F1$  being set to 0 if $P+R=0$ as usual.

\section{Compute requirements}\label{app:compute}

We run our \model evaluations on Ubuntu servers each with 8 NVIDIA RTX A6000 GPUs. As mentioned previously, we do not train any models and make use of the OpenAI and together.ai APIs to perform inference using LLMs. This means we do not require (local) GPUs for LLM inference but there are many instances when \model may benefit from having access to a GPU (\eg, for image inpainting). While inference time varies per benchmark, running \mnm evaluations (200 queries) with 24 parallel processes on a single 8 GPU server takes approximately ${\sim}$12 minutes in wall clock time (96 minutes of GPU time).

\section{Library and Tool Descriptions}\label{app:prompts}

Here we provide the library and tool descriptions used in our case studies as well as quantitative evaluation -
\begin{itemize}
    \item Figures~\ref{fig:lib_desc_codenav},~\ref{fig:lib_desc_mnm},~\ref{fig:lib_desc_phi} shows the library descriptions used by \model in the three case studies.
    \item Figures~\ref{fig:lib_desc_mnm},~\ref{fig:lib_desc_mthreeeval},~\ref{fig:lib_desc_apibank} show the library descriptions used by \model for quantiative evaluation on \mnm, \mthreeeval, and \apibank respectively.
    \item Figures~\ref{fig:tool_desc_mnm},~\ref{fig:tool_desc_mthreeeval},~\ref{fig:tool_desc_apibank} show the tool descriptions used for the tool-use baselines for \mnm, \mthreeeval, and \apibank respectively. Note that the tool descriptions are significantly more detailed than the corresponding library descriptions.
\end{itemize}

\section{Retrieval Response Example}\label{app:retr_response_example}

\begin{figure}[t]
\centering
\includegraphics[width=0.5\textwidth]{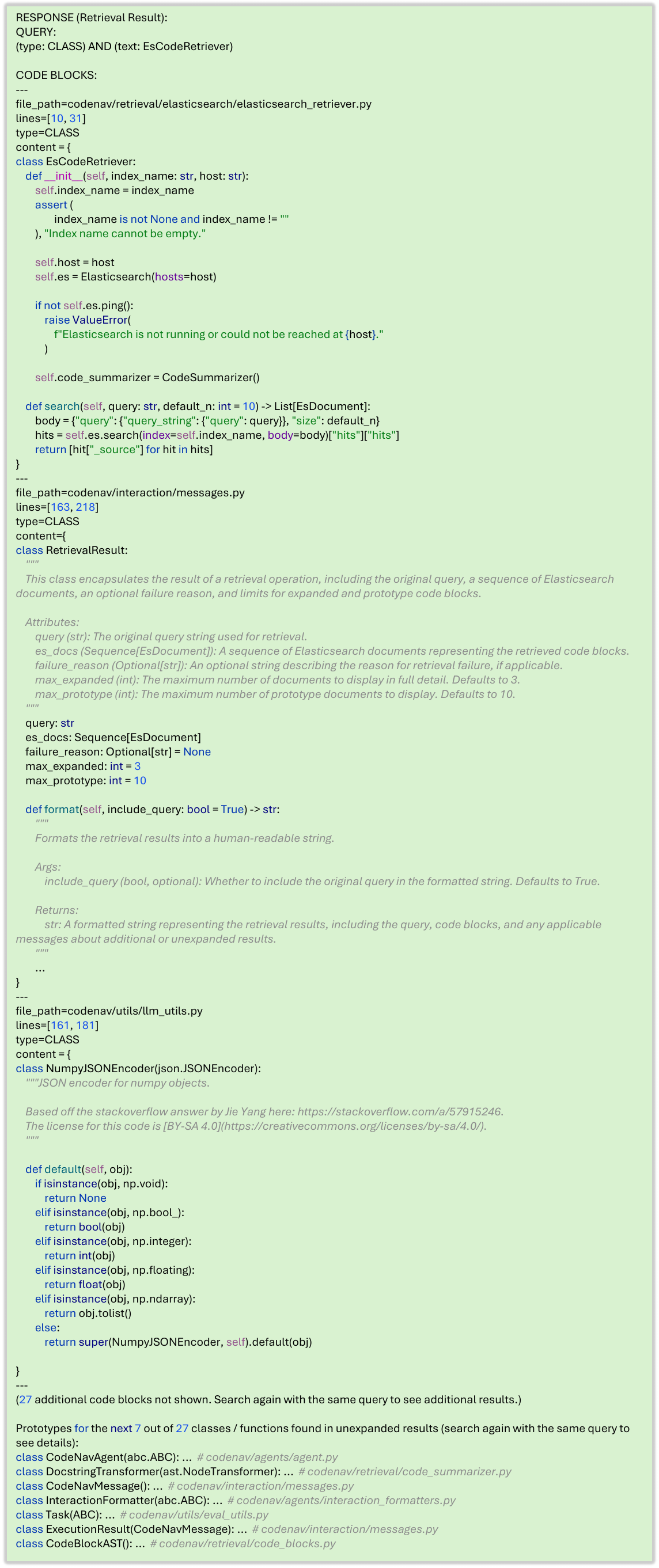}
\caption{\textbf{Full retrieval response example.} An example of a full retrieval. This corresponds to an expanded version of \hlresbg{R5} in Fig.~\ref{fig:codenav-on-codenav}.}
\label{fig:retrieval_result}
\end{figure}

We show a full example of a retrieval response in Fig.~\ref{fig:codenav-on-codenav}. This corresponds to an expanded version of \hlresbg{R5} in Fig.~\ref{fig:codenav-on-codenav} where we only showed one response for brevity. Notice: (1) the collection of class and function prototypes/signatures shown at the bottom of the retrieval response and (2) that the 3 expanded retrieved results contain a mix of full code (for the \texttt{EsCodeRetriever} and \texttt{NumpyJSONEncoder} classes) as well as automatically generated summary docstrings (for the \texttt{RetrievalResult} class).

\section{Limitations}\label{sec:limitations}

While \model is capable of some producing impressive results, we highlight three key limitations. (1) Our current implementation of \model assumes that the agent produces Python code (and that the given codebase is a Python codebase). While extending to use other languages is not a significant engineering challenge, it is possible that LLMs' performance degrades as one moves away from popular languages like Python, for which there is significant data on the web. (2) Our agent has no long-term memory that is available across queries. This means that, given the same query twice, \model will make the same errors and repeat the same searches. (3) Finally, \model performance is strongly dependent on the underlying LLM and performance can sharply degrade when using smaller LLMs. This means that most researchers can only use \model through the use of paid APIs and larger-scale experimentation when using these APIs can be costly.

\section{Societal impact}\label{app:societal-impact}

While \model is not unique in this respect, the growing popularity of increasingly competent LLM agents has the potential to automate or augment many skilled tasks. While automation has brought about many societal positives on aggregate, it can clearly has a profoundly negative impact on anyone whose may lose their job. The environmental impact (via the huge energy costs of running these LLMs) may also be significant.

As a more acute potential negative impact: in the code-use paradigm, the underlying agent the \model agent has the ability to run arbitrary code on the user's machine. Given this, it can be dangerous unintentionally (there is nothing stopping \model from making a logic error, \eg filtering a file list incorrectly and then accidentally deleting all files on the computer) and intentionally (\eg, a malicious party might upload tainted model weights, or intercept API calls, so as to run a ``randomsome ware'' scam when executed by a \model agent). These dangers are somewhat easier to mitigate in the tool-use paradigm as, when constrained to use only certain tools, there are fundamentally fewer attack vectors to consider.

\begin{figure}[h]
    \begin{tcolorbox}[colback=lightgray!25, colframe=white, width=\textwidth]
        \scriptsize
        The codebase you will use is called `codenav`. It is a library for creating LLM agents that can interact with one of the avilable environments to solve the user queries that require using an external codebase. For example, an agent can interact with "PythonCodeEnv" for code execution, and with "RetrievalEnv" for retrieving code snippets from an ElasticSearch index. It provides an "Episode" class for running this interaction with a specific agent and a set of environments. Here's the directory structure:
        \\

codenav/agents/ - contains subdirectories that store implementations of LLM agents implemented with different LLMs

codenav/environments/ - contains environments that the agent can interact with

codenav/interaction/ - contains Episode and messages implementations (messages is how agent interacts with environments)

codenav/prompts/ - stores various system prompts for the LLM agent

codenav/retrieval/ - various files related to creating elastic search index and retrieving items from the index

codenav/utils/ - contains various utility python files

codenav/constants.py - contains important constants
    \end{tcolorbox}
    \caption{\textbf{Library description for \model case study in Sec.~\ref{sec:codenav_case_study}}}
    \label{fig:lib_desc_codenav}
\end{figure}

\begin{figure}[h]
    \begin{tcolorbox}[colback=lightgray!25, colframe=white, width=\textwidth]
        \scriptsize
        The codebase you'll be using to solve user tasks is called `mnm`. It has a single tool\_api.py file which contains functions for various image, text, and audio related tasks. Specifically, here's a high-level summary of available functions:
\\

- text understanding functions: for tasks like text generation, summarization, classification, answering questions based on a text context
\\

- image understanding functions: for tasks like image classification (1000 IMAGENET categories only), image captioning, answering questions about image (use this if you have a question that can't be answered with IMAGENET classification), detecting objects (producing bounding boxes and labels for COCO categories) and segmenting objects (producing segmentation masks and labels for COCO categories), transcribing alphanumeric characters in an image to text (also known as optical character recognition).
\\

- image editing functions: for generating images give a text description, editing images given the original image and a description of how to modify the image (can handle queries that require replacing or removing certain objects in the scene without detecting the object first), image cropping, or achieving effects like color pop and background blur given the segmented objects which you want to highlight in the image.
\\

- information retrieval functions: for retrieving factual information or interesting facts about dates, years, numbers, movies, weather, geographical coordinates of a city, wikipedia articles, or fun trivia. Also includes a love calculator for checking compatibility given two names.
\\

- object centric functions: these are functions that accept a list of detected or segmented objects for tasks like counting, selecting an object, tagging an image with objects (drawing bounding boxes and labels), or replacing objects with emojis. Note that these functions here do not detect the objects themselves.
\\

- audio understanding functions: for tasks related to audio understanding like speech recognition
    \end{tcolorbox}
    \caption{\textbf{Library description for \mnm as well as the multimodal case study in Sec.~\ref{sec:mnm_case_study}}}
    \label{fig:lib_desc_mnm}
\end{figure}

\begin{figure}[h]
    \begin{tcolorbox}[colback=lightgray!25, colframe=white, width=\textwidth]
        \scriptsize
        The codebase you will use is called `phi` (already installed via pip).\\

Here are some key file paths:

phi/tools/email.py - for sending emails (requires passkey which can be read from environment variable GOOGLE\_KEY; use ABC as sender name and ABC@DEF.org as sender email id)

phi/tools/arxiv\_toolkit.py - for search arxiv for research papers

phi/tools/tavily.py - an AI driven search engine (requires pass key stored in environment variable TAVILY\_KEY)

phi/tools/duckduckgo.py - a search engine similar to google 

phi/tools/newspaper4k.py - for scarping news articles

phi/tools/pubmed.py - for searching biomedical and life sciences literature

phi/tools/yfinance.py - for fetching financial data like stock prices from Yahoo Finance

phi/tools/wikipedia.py - for searching wikipedia (need to use json.loads on the output to extract content)
    \end{tcolorbox}
    \caption{\textbf{Library description for \phidata case study in Sec.~\ref{sec:phidata_case_study}}}
    \label{fig:lib_desc_phi}
\end{figure}

\begin{figure}[h]
    \begin{tcolorbox}[colback=lightgray!25, colframe=white, width=\textwidth]
        \scriptsize
        The codebase you will use is called `m3eval`. It has the following directory structure:\\

m3eval/travel\_planner.py - function for planning travel including finding flights, making hotel reservation, and budget calculations

m3eval/browser.py - contains the WebBrowser class for navigating web pages

m3eval/dna\_sequencer.py - various functions related to dna sequencing

m3eval/message\_decoder.py - functions for encoding and decoding messages including converting hex string to ascii and decoding caesar cipher

m3eval/trade\_calculator.py - functions for currency conversion, calculating tariffs etc
    \end{tcolorbox}
    \caption{\textbf{Library description for \mthreeeval.}}
    \label{fig:lib_desc_mthreeeval}
\end{figure}

\begin{figure}[htbp]
    \begin{tcolorbox}[colback=lightgray!25, colframe=white, width=\textwidth]
        \scriptsize
        The codebase you'll be using to solve user tasks is called `api-bank`. This codebase implements a collection of different APIs as classes, here is a full list of these APIs:\\

AddAgenda: The API for adding a agenda item includes content, time and location.

AddAlarm: The API for setting an alarm includes a parameter for the alarm time.

AddMeeting: This API allows users to make a reservation for a meeting and store the meeting information (e.g., topic, time, location, attendees) in the database.

AddReminder: The API for adding a reminder item includes content and time.

AddScene: This API adds a scene of smart home system, given the scene name and a list of smart devices

AppointmentRegistration: This API registers an appointment of hospital.

BookHotel: This API orders a hotel room. Two rooms are ordered if the number of adults is greater than 2. Only one order can be made at same time.

Calculator: This API provides basic arithmetic operations: addition, subtraction, multiplication, and division.

CancelRegistration: This API cancels the registration of a patient given appointment ID.

CancelTimedSwitch: Cancels a timed switch for a smart device.

CheckToken: Check the user token.

DeleteAccount: Delete an account.

DeleteAgenda: The API for deleting a schedule item includes parameters for token, content, time, and location.

DeleteAlarm: The API for removing an alarm includes a parameter for the time.

DeleteMeeting: This API allows users to delete a reservation for a meeting and remove the meeting information in the database.

DeleteReminder: The API for deleting a reminder item includes content and time.

DeleteScene: This API deletes a scene by its name.

Dictionary: This API searches the dictionary for a given keyword.

DocumentQA: This API answers the question from a given document url.

EmergencyKnowledge: This API searches for a given symptom for emergency knowledge.

ForgotPassword: Sends an email to the user with a link to reset the password. Need call twice, first with 'Forgot Password' status to get the verification code, then call again with 'Verification Code' status to change the password. Must pass the name of the parameters when calling the API, like ForgotPassword(status='Forgot Password', username='username').

GetToday: This API gets the current date.

GetUserToken: Get the user token by username and password.

ImageCaption: This API generates a caption for a given image.

ModifyAgenda: The API for modifying a schedule item includes parameters for content, time, and location.

ModifyAlarm: The API for modifying an alarm includes a parameter for the from\_time to to\_time.

ModifyMeeting: This API allows users to modify a reservation for a meeting

ModifyPassword: The API for modifying the password of the account.

ModifyRegistration: This API modifies the registration of a patient given appointment ID.

ModifyReminder: The API for deleting a reminder item includes content and time.

ModifyScene: This API modifies a scene of smart home system, given the scene name and a list of smart devices

OpenBankAccount: This is an API for opening a bank account for a user, given the account, password and name.

PlayMusic: This API triggers a music player to play music.

QueryAgenda: The API for getting a schedule item includes parameters for token, content, time, and location.

QueryAlarm: The API for querying alarm clock, help user to check the alarm clock they have set.

QueryBalance: This API queries the balance of a given user.

QueryHealthData: This API queries the recorded health data in database of a given user and time span.

QueryHistoryToday: This API queries the history of the given date.

QueryMeeting: This API allows users to query the information of a meeting.

QueryRegistration: This API queries the registration of a patient, given patient ID.

QueryReminder: The API for querying a reminder item includes content and time.

QueryScene: This API queries a scene of smart home system, given the scene name

QueryStock: This API queries the stock price of a given stock code and date.

RecordHealthData: This API records the health data of a user.

RegisterUser: The API for registering a account, given the username, password and email.

SearchEngine: This API searches for a given keyword for search engine.

SendEmail: This API for sending email, given the receiver, subject and content.

SpeechRecognition: This API recognizes the speech from a given audio url.

SymptomSearch: This API searches for a given symptom.

TimedSwitch: This API for setting a timed switch for a smart device.

ToolSearcher: Searches for relevant tools in library based on the keywords.

Translate: Translate the text to the target language.

Wiki: This API for searching a keyword in Wikipedia.
    \end{tcolorbox}
    \caption{\textbf{Library description for \apibank.}}
    \label{fig:lib_desc_apibank}
\end{figure}

\begin{figure}[h]
    \begin{tcolorbox}[colback=lightgray!25, colframe=white, width=\textwidth]
        \scriptsize
        The codebase you'll be using to solve user tasks is called `mnm`, this codebase consists of a single file called tool\_api.py which contains the following functions:\\

text\_generation(text) -> text: It takes an input text prompt and outputs a text that is most likely to follow the input text.

text\_summarization(text) -> text: it takes a paragraph of text and summarizes into a few sentences.

text\_classification(text) -> text: It takes a text and classifies it into a category in the model's vocaburary (e.g. positive or negative based on its sentiment).

question\_answering(text, question) -> text: It takes a text and a question, and outputs an answer to that question based on the text.

image\_generation(text) -> image: It takes a text prompt and generates an image that matches the text description.

image\_captioning(image) -> text: It takes an image and generates a text caption of the image.

optical\_character\_recognition(image) -> text: It takes an image and outputs recognized texts in the image.

image\_classification(image) -> text: It takes an image and classifies the subject in the image into a category such as cat or dog.

image\_editing(image, prompt) -> image: It takes an image and a text prompt and outputs a new image based on the text.

object\_detection(image) -> image, objects: It takes an image and outputs rectangular bounding boxes of objects detected in the image.

image\_segmentation(image) -> image, objects: It takes an image, segments it into different parts, and outputs segmentation masks of any shape for the parts.

automatic\_speech\_recognition(audio) -> text: It takes an audio file and produces a transcription of the audio.

visual\_question\_answering(image, question) -> text: It takes an image and a question about the image, and generates an answer to the question.

image\_crop(image, object) -> image: It takes an image and 4 numbers representing the coordinates of a bounding box and crops the image to the region within the box.

image\_crop\_left(image) -> image: It takes an image, crops and keeps the left part of the image.

image\_crop\_right(image) -> image: It takes an image, crops and keeps the right part of the image.

image\_crop\_top(image) -> image: It takes an image, crops and keeps the top part of the image.

image\_crop\_bottom(image) -> image: It takes an image, crops and keeps the bottom part of the image.

background\_blur(image, object) -> image: It takes an image and one or multiple objects in the foreground, and returns an image where the backgroud is blurred.

color\_pop(image, object) -> image: It takes an image and one or multiple objects, and returns an image where only the object is colored and the rest is black and white.

count(objects) -> number: It takes a list of objects and returns the count of the objects.

tag(image, objects) -> image: It takes an image and a list of objects with their bounding boxes and classes, and tags all the objects.

select\_object(objects, object\_name) -> object: It takes a list of objects, and selects the object based on the input object name.

emoji(image, object, emoji) -> image: It takes an image and the bounding box coordinates of one or multiple objects, and replaces the object with an emoji (e.g. angry/flushed/crying/dizzy/sleepy/grimacing/kissing/smiling\_face, alien, ghost, goblin etc).

get\_date\_fact(date) -> text: It provides interesting facts about dates.

get\_year\_fact(year) -> text: It provides interesting facts about years.

get\_math\_fact(number) -> text: It provides interesting math facts about numbers.

get\_trivia\_fact(number) -> text: It provides interesting trivia facts about number.

love\_calculator(first\_name, second\_name) -> number: Enter your name and the name of your partner/lover/crush to find Love compatibility \& chances of successful love relationship.

get\_location(city) -> lon, lat: Convert a city name or address to geographical coordinates using OpenStreetMap's Nominatim API.

search\_movie(movie\_title, movie\_year) -> text: Retrieve basic movie information, including title, year, genre, and director.

get\_weather(lon, lat) -> objects: Provides weather forecast data based on specific geographical coordinates.

wikipedia\_simple\_search(text) -> text: Perform a basic search query on Wikipedia to retrieve a summary of the most relevant page.\\

NOTE - all of the above functions produce a python dictionary as output. The function signatures above show the key present in the dictionary. \\ 

For example, get\_year\_fact(year) -> text means that the fact can be accessed through \\
output = get\_year\_fact(year) \\
fact = output['text']
    \end{tcolorbox}
    \caption{\textbf{Tool description for \mnm.}}
    \label{fig:tool_desc_mnm}
\end{figure}

\begin{figure}[htbp]
    \begin{tcolorbox}[colback=lightgray!25, colframe=white, width=\textwidth]
        \scriptsize
        The codebase you will use is called `m3eval`. It has the following files and functions implemented in those files\\

file: m3eval/message\_decoder.py

[1] convert\_hex\_to\_ascii: Converts a hexadecimal string to ASCII. Arguments: hex\_string (str)
    Signature: convert\_hex\_to\_ascii(hex\_string: str) -> str
    
[2] reverse\_string: Reverses a string. Arguments: string (str)
    Signature: reverse\_string(string: str) -> str
    
[3] caesar\_decode: Decodes a string using the Caesar cipher. Arguments: message (str), shift (int)
    Signature: caesar\_decode(message: str, shift: int) -> str
    
[4] string\_length: Finds the length of a string. Arguments: string (str)
    Signature: string\_length(string: str) -> int

[5] minimum\_value: Finds the minimum value from given arguments. Arguments: *args (variable number of arguments)
    Signature: minimum\_value(*args) -> int/float

[6] maximum\_value: Finds the maximum value from given arguments. Arguments: *args (variable number of arguments)
    Signature: maximum\_value(*args) -> int/float
\\

file: m3eval/dna\_sequencer.py

[7] count\_nucleotides: Counts the occurrences of each nucleotide in a DNA sequence. Arguments: dna\_sequence (str)
    Signature: count\_nucleotides(dna\_sequence: str) -> dict

[8] transcribe\_dna\_to\_mrna: Transcribes DNA sequence to mRNA. Arguments: dna\_sequence (str)
    Signature: transcribe\_dna\_to\_mrna(dna\_sequence: str) -> str

[9] translate\_mrna\_to\_amino\_acid: Translates mRNA sequence to a chain of amino acids. Arguments: mrna\_sequence (str)
    Signature: translate\_mrna\_to\_amino\_acid(mrna\_sequence: str) -> str

[10] find\_max\_nucleotide: Return the nucleotide (str) with the maximum count (int). Arguments: nucleotide\_counts in the form of (k1, v1, k2, v2, ..., kn, vn)
    Signature: find\_max\_nucleotide(*args) -> (str, int)

[11] is\_valid\_dna\_sequence: Checks if the DNA sequence is valid. Arguments: dna\_sequence (str)
    Signature: is\_valid\_dna\_sequence(dna\_sequence: str) -> bool

[12] reverse\_transcribe\_mrna\_to\_dna: Reverse transcribes mRNA sequence to DNA. Arguments: mrna\_sequence (str)
    Signature: reverse\_transcribe\_mrna\_to\_dna(mrna\_sequence: str) -> str
\\

file: m3eval/trade\_calculator.py

[13] convert\_currency: Converts the commodity price to local currency. Arguments: base\_price (float), conversion\_rate (float)
    Signature: convert\_currency(base\_price: float, conversion\_rate: float) -> float

[14] calculate\_tariff: Calculates the trade tariff based on the converted price. Arguments: price (float), tariff\_rate (float, in %
    Signature: calculate\_tariff(price: float, tariff\_rate: float) -> float

[15] estimate\_final\_value: Estimates the final trade value including the tariff. Arguments: price (float), tariff (float)
    Signature: estimate\_final\_value(price: float, tariff: float) -> float

[16] calculator: Evaluates the given expression and returns the result. Accepts a calculation expression as input. For example, "2 + (3 * 4)" will return 14.
    Signature: calculator(expression: str) -> float

[17] find\_minimum: Finds the minimum value among the given arguments. Accepts variable number of float arguments.
    Signature: find\_minimum(*args: float) -> float

[18] find\_maximum: Finds the maximum value among the given arguments. Accepts variable number of float arguments.
    Signature: find\_maximum(*args: float) -> float
\\

file: m3eval/travel\_planner.py

[19] find\_flights: Finds flights based on source, destination and date. Arguments: from\_location (str), to\_location (str), date (str) in YYYY-MM-DD format.
Returns a list of flights, each represented as a dictionary with keys "from\_location", "to\_location" (destination), "date", and "price".
Example: [{"from\_location": "A", "to\_location": "B", "date": "2023-12-25", "price": 450}]
    Signature: find\_flights(destination: str, date: str) -> List[Dict]

[20] book\_hotel: Books a hotel based on location and preferences. Arguments: location (str), *preferences (variable number of str arguments).
Returns a list of hotels, each represented as a dictionary with keys "location", "preferences", "price\_per\_night", and "rating".
Example: [{"location": "A", "preferences": ["wifi", "pool"], "price\_per\_night": 120, "rating": 4}]
    Signature: book\_hotel(location: str, *preferences: str) -> List[Dict]

[21] budget\_calculator: Calculates the total budget for a trip. Arguments: flight\_price (float), hotel\_price\_per\_night (float), num\_nights (int).
Returns the total budget (float).
    Signature: budget\_calculator(flight\_price: float, hotel\_price\_per\_night: float, num\_nights: int) -> float
\\

file: m3eval/browser.py 

Note - To use the browser functions first create a browser instance using browser=WebBrowser()

[22] browser.click\_url: Clicks on a URL. A clickable URL looks like [Clickable '<url\_argument>'] in the webpage.
Arguments: url (str).
Returns the rendered content of the webpage after clicking the URL showing on the current rendered page.
    Signature: browser.click\_url(url: str) -> str

[23] browser.go\_to\_previous\_page(): Goes back to the previous page. It has no arguments.
After going back to the previous page, return the rendered content of the webpage.
    Signature: browser.go\_to\_previous\_page() -> str

[24] browser.scroll\_down: Scrolls down the view. It has no arguments.
Returns the rendered content of the webpage after scrolling down.
    Signature: browser.scroll\_down() -> str

[25] browser.scroll\_up: Scrolls up the view. It has no arguments.
Returns the rendered content of the webpage after scrolling up.
    Signature: browser.scroll\_up() -> str

[26] browser.view: Return the current view in string format of the rendered webpage. It has no arguments.
Returns the rendered content of the webpage.
You should call this when you want to see the rendered content of the current webpage.
    Signature: browser.view() -> str
    \end{tcolorbox}
    \caption{\textbf{Tool description for \mthreeeval.}}
    \label{fig:tool_desc_mthreeeval}
\end{figure}

\begin{figure}[htbp]
    \begin{tcolorbox}[colback=lightgray!25, colframe=white, width=\textwidth]
        \scriptsize
        The codebase you'll be using to solve user tasks is called `api-bank`. This codebase implements a collection of different APIs as classes, here is a full list of these APIs:

AddAgenda().call(token, content, time, location)
Import as: from apis import AddAgenda
Description: The API for adding a agenda item includes content, time and location.
Arguments:
- token (str): User's token.
- content (str): The content of the agenda.
- time (str): The time for agenda. Format: %
- location (str): The location of the agenda.
Returns: A dictionary whose "output" key has value of type str and description success or failed.

AddAlarm().call(token, time)
Import as: from apis import AddAlarm
Description: The API for setting an alarm includes a parameter for the alarm time.
Arguments:
- token (str): User's token.
- time (str): The time for alarm. Format: %
Returns: A dictionary whose "output" key has value of type str and description success or failed.

AddMeeting().call(token, meeting\_topic, start\_time, end\_time, location, attendees)
Import as: from apis import AddMeeting
Description: This API allows users to make a reservation for a meeting and store the meeting information (e.g., topic, time, location, attendees) in the database.
Arguments:
- token (str): User's token.
- meeting\_topic (str): The title of the meeting, no more than 50 characters.
- start\_time (str): The start time of the meeting, in the pattern of %
- end\_time (str): The end time of the meeting, in the pattern of %
- location (str): The location where the meeting to be held, no more than 100 characters.
- attendees (list(str)): The attendees of the meeting, including names, positions and other information.
Returns: A dictionary whose "output" key has value of type str and description success or failed.

AddReminder().call(token, content, time)
Import as: from apis import AddReminder
Description: The API for adding a reminder item includes content and time.
Arguments:
- token (str): User's token.
- content (str): The content of the conference.
- time (str): The time for conference. Format: %
Returns: A dictionary whose "output" key has value of type str and description success or failed.

AddScene().call(name, devices)
Import as: from apis import AddScene
Description: This API adds a scene of smart home system, given the scene name and a list of smart devices
Arguments:
- name (str): The name of the scene.
- devices (list): The list of smart devices, containing the name and description. Format be like [{"name": "light", "description": "Smart light in the kitchen"}, {"name": "oven", "description": "Smart oven in the kitchen"}, {"name": "range hood", "description": "Smart range hood in the kitchen"}]
Returns: A dictionary whose "output" key has value of type str and description Whether succeed..

AppointmentRegistration().call(patient\_name, date, doctor\_name)
Import as: from apis import AppointmentRegistration
Description: This API registers an appointment of hospital.
Arguments:
- patient\_name (str): The name of patient.
- date (str): The date of appointment. Format be like %
- doctor\_name (str): The name of appointed doctor.
Returns: A dictionary whose "output" key has value of type str and description The ID of appointment..

BookHotel().call(hotel\_name, check\_in\_time, check\_out\_time, room\_count, adult\_count, child\_count)
Import as: from apis import BookHotel
Description: This API orders a hotel room. Two rooms are ordered if the number of adults is greater than 2. Only one order can be made at same time.
Arguments:
- hotel\_name (str): The name of the hotel.
- check\_in\_time (str): The time to check in. Format: %
- check\_out\_time (str): The time to check out. Format: %
- room\_count (int): The number of rooms to order.
- adult\_count (int): The number of adults.
- child\_count (int): The number of children.
Returns: A dictionary whose "output" key has value of type str and description The ID of the order..

Calculator().call(formula)
Import as: from apis import Calculator
Description: This API provides basic arithmetic operations: addition, subtraction, multiplication, and division.
Arguments:
- formula (str): The formula that needs to be calculated. Only integers are supported. Valid operators are +, -, *, /, and (, ). For example, '(1 + 2) * 3'.
Returns: A dictionary whose "output" key has value of type float and description The result of the formula..

CancelRegistration().call(appointment\_id)
Import as: from apis import CancelRegistration
Description: This API cancels the registration of a patient given appointment ID.
Arguments:
- appointment\_id (str): The ID of appointment.
Returns: A dictionary whose "output" key has value of type str and description The status of cancellation..\\

.

.

.
\\

SearchEngine().call(keyword)
Import as: from apis import SearchEngine
Description: This API searches for a given keyword for search engine.
Arguments:
- keyword (str): The keyword to search.
Returns: A dictionary whose "output" key has value of type list and description The list of results..

SendEmail().call(receiver, subject, content)
Import as: from apis import SendEmail
Description: This API for sending email, given the receiver, subject and content.
Arguments:
- receiver (str): The receiver address of the email.
- subject (str): The subject address of the email.
- content (str): The content of the email.
Returns: A dictionary whose "output" key has value of type str and description The status of the email..

SpeechRecognition().call(url)
Import as: from apis import SpeechRecognition
Description: This API recognizes the speech from a given audio url.
Arguments:
- url (str): The url to download the audio. It should end with .wav.
Returns: A dictionary whose "output" key has value of type str and description The transcript of the audio..

SymptomSearch().call(symptom)
Import as: from apis import SymptomSearch
Description: This API searches for a given symptom.
Arguments:
- symptom (str): The symptom to search.
Returns: A dictionary whose "output" key has value of type list and description The list of results. Format be like [{"name":possible disease name, "description": disease details},...].

TimedSwitch().call(name, time, on)
Import as: from apis import TimedSwitch
Description: This API for setting a timed switch for a smart device.
Arguments:
- name (str): The name of the smart device.
- time (str): The time to switch the device on or off. Format: %
- on (bool): Whether to switch the device on or off.
Returns: A dictionary whose "output" key has value of type str and description Whether the time switch is successful..

ToolSearcher().call(keywords)
Import as: from apis import ToolSearcher
Description: Searches for relevant tools in library based on the keywords.
Arguments:
- keywords (str): The keyword to search for.
Returns: A dictionary whose "output" key has value of type Union[List[dict], dict] and description The best match tool(s)..

Translate().call(src, src\_lang, tgt\_lang)
Import as: from apis import Translate
Description: Translate the text to the target language.
Arguments:
- src (str): The text to be translated.
- src\_lang (str): [Optional] The source language to translate from. Default is auto.
- tgt\_lang (str): [Optional] The target language to translate to. Default is english/en.
Returns: A dictionary whose "output" key has value of type str and description The translated text..

Wiki().call(keyword)
Import as: from apis import Wiki
Description: This API for searching a keyword in Wikipedia.
Arguments:
- keyword (str): The keyword to search.
Returns: A dictionary whose "output" key has value of type dict and description The list of results. Format be like {"url": "xxx", "summary": "xxx", "content": "xxx"}.
        
    \end{tcolorbox}
    \caption{\textbf{Tool description for \apibank.} We only show the beginning and end of the full description for brevity.}
    \label{fig:tool_desc_apibank}
\end{figure}

\clearpage
\newpage
\clearpage
\newpage
\clearpage
\newpage

\end{document}